\documentclass[runningheads]{llncs}
\usepackage{graphicx}
\usepackage{kotex}
\usepackage{comment}
\usepackage{subfig}
\usepackage{makecell}
\usepackage{xcolor}
\usepackage{textcomp}
\usepackage{algorithm}          
\usepackage{algpseudocode}      
\usepackage{xspace}
\usepackage{array}

\usepackage{subfig}
\usepackage{amsmath}
\usepackage{amssymb}
\usepackage{mathtools}
\usepackage{multirow}
\usepackage{booktabs}
\usepackage{multicol,multirow}

\usepackage{cite}

\begin{document}
\title{Improving Real Estate Appraisal with POI Integration and Areal Embedding\\ \large Preprint}
\titlerunning{Improving Real Estate Appraisal with POI and Areal Embedding}


%
%



\author{Sumin Han\inst{1} \and Youngjun Park\inst{1} \and Sonia Sabir\inst{1} \and Jisun An\inst{2} \and Dongman Lee\inst{1}}
\institute{%
  \department{School of Computing}
  \institution{Korea Advanced Institute of Science and Technology}
  \streetaddress{291 Daehak-ro, Yuseong-gu}
  \city{Daejeon}
  \country{Republic of Korea}
  \postcode{34141}
}
\institute{Korea Advanced Institute of Science and Technology (KAIST)\email{\{hsm6911,youngjourpark,sonia,dlee\}@kaist.ac.kr} \and Indiana University Bloomington \email{jisunan@iu.edu}}

%
%
%
\maketitle              
\begin{abstract}


Despite advancements in real estate appraisal methods, this study primarily focuses on two pivotal challenges. Firstly, we explore the often-underestimated impact of Points of Interest (POI) on property values, emphasizing the necessity for a comprehensive, data-driven approach to feature selection. Secondly, we integrate road-network-based Areal Embedding to enhance spatial understanding for real estate appraisal. We first propose a revised method for POI feature extraction, and discuss the impact of each POI for house price appraisal. Then we present the Areal embedding-enabled Masked Multihead Attention-based Spatial Interpolation for House Price Prediction (AMMASI) model, an improvement upon the existing ASI model, which leverages masked multi-head attention on geographic neighbor houses and similar-featured houses. Our model outperforms current baselines and also offers promising avenues for future optimization in real estate appraisal methodologies.

\end{abstract}

\section{Introduction}

Real estate appraisers play a crucial role in delivering essential property valuations that serve diverse purposes, including buying and selling, securing financing, determining taxation, and establishing insurance coverage.
Conventional real estate appraisal often relies on subjective methods such as the sales comparison approach, where a property's value is assessed by comparing it to similar properties in the same market. 
While the Hedonic Price Model~\cite{rosen1974hedonic} remains foundational in real estate appraisal, positing that a property's value is shaped by its individual characteristics, the inherent subjectivity and context dependency in establishing comparability makes property valuation a complex task. 
Recently, there has been a surge in deep learning-based methodologies, aiming to determine land prices that reflect spatial understanding and correlations among houses in large-scale datasets\cite{ASI, MugRep, ReGram, STRAP, DoRA}.

Despite the significant strides made in real estate appraisal methods, we mainly focus on two major challenges in this paper:

\begin{enumerate}
    \item \textbf{POI Integration:} While numerous studies have focused on physical house feature selection in property valuation models, a common issue is the lack of explicit identification of which POI is truly pivotal in determining property values~\cite{de2018economic}. This challenge highlights the need for a more comprehensive and data-driven approach to feature selection, which can help pinpoint the most influential factors in property valuation.

    \item \textbf{Areal Embedding:} The recent study demonstrates that employing trainable regional embedding can improve real estate appraisal by incorporating comprehensive spatial knowledge beyond latitude and longitude~\cite{LESR}. However, these methods often demand sophisticated and comprehensive data, such as mobility data or satellite images, posing a challenge for the integration of scalable areal embedding techniques.

\end{enumerate}

To address these challenges, we employ a simplified method for POI feature extraction and introduce the Areal Embedding-based Masked Multihead Attention-based Spatial Interpolation for House Price Prediction (AMMASI) model as a refinement of the existing ASI~\cite{ASI} model. Applying masked multi-head attention to both geographic neighbor houses and those with similar features, our model outperforms current baselines. We summarize our key contributions as follows:

\begin{itemize}
    \item[$\bullet$] We introduce a novel method named AMMASI, building upon the foundation of the ASI model, with publicly available code implementation \footnote{\url{https://anonymous.4open.science/r/AMMASI-C625}}.
    \item[$\bullet$] We propose integrating Point of Interest (POI) data with publicly accessible OpenStreetMap by engineering features with Gaussian proximity.
    \item[$\bullet$] Our model utilizes road network-based Areal Embedding to incorporate geographical insights, presenting a simplified approach in comparison to existing methods.
\end{itemize}



\section{Related Work}

\subsection{Recent Deep Neural Networks on Real Estate Appraisal}

Recent advancements in real estate appraisal have placed a strong emphasis on innovative approaches such as the selection of reference houses or community construction. PDVM\cite{PDVM} employs K-nearest similar house sampling for sequence generation, while ASI\cite{ASI} focuses on estimating house prices through a hybrid attention mechanism. MugRep\cite{MugRep} incorporates a hierarchical heterogeneous community graph convolution module. ReGram\cite{ReGram} utilizes a neighbor relation graph with an attention mechanism, and ST-RAP\cite{STRAP} adopts a hierarchical model, integrating temporal and spatial aspects alongside amenities. Additionally, DoRA\cite{DoRA} introduces a domain-based self-supervised learning framework with pretraining on geographic prediction. Collectively, these methodologies comprehensively address the complexities of real estate appraisal by considering spatial, temporal, and community factors.

\subsection{Using POI Features for Real Estate Appraisal}

The utilization of Points of Interest (POI) in real estate appraisal has a longstanding history dating back to the hedonic model. These attributes, encompassing schools, parks, shopping centers, and transportation hubs, exert diverse impacts on property values, posing a challenge in quantifying their relative significance. Ottensmann et al.~\cite{ottensmann2008urban} introduced the use of distance to the Central Business District (CBD) as an additional house attribute, while Xiao et al.~\cite{xiao2017beijingPOI} proposed accessibility indices using a distance-based metric ($1 - d_{ij}/D$) to measure proximity to amenities. ASI~\cite{ASI} took a different approach by leveraging the number of POIs in each dataset through the crawling of external APIs, although their POI dataset remains non-public. On the other hand, DoRA~\cite{DoRA} adopted a tabular format, distinguishing between Yes In My Back Yard (YIMBY) facilities, like parks and schools, and Not In My Back Yard (NIMBY) facilities, such as power stations and landfills. The calculation of the number of POIs for the real estate property was performed using the Euclidean distance.

\subsection{Areal Embedding}

Addressing the Learning an Embedding Space for Regions (LESR) problem, RegionEncoder\cite{LESR} presents a holistic approach to jointly learning vector representations for discrete urban regions, leveraging diverse data sources such as satellite images, point-of-interest data, human mobility patterns, and spatial graphs. Hex2vec~\cite{Hex2vec} introduces an innovative technique for learning vector representations of OpenStreetMap (OSM) regions, incorporating considerations of urban functions and land use within a micro-region grid. Shifting focus to Representation Learning for Road Networks (RLRN), RN2Vec~\cite{road2vec} leverages the neural network model to obtain embeddings of intersections and road segments. Node2Vec~\cite{node2vec} is often referenced as a baseline for graph-based node representation learning in this domain.

\section{Problem Formulation}

Let $\mathbf{f}_i \in \mathbb{R}^{N_f}$ represent the physical features of house $i$, such as the number of beds and rooms. Longitude and latitude are denoted as $loc^x_i$ and $loc^y_i$, respectively, and are not included in the house features.
Furthermore, we define global geometric features of Points of Interest (POIs) and roads, that every house can leverage.
Polygon geometries are defined for $N_P$ types of Points of Interest (POIs), such as commercial land uses and parks. Each POI type has corresponding regions or buildings represented as polygon geometries, and the union geometry for each POI type is denoted as $GEO^{POI}_{1, ..., N_P}$.
Line geometries of roads are given and denoted as $GEO^{ROAD}_{1, ..., N_R}$.
The goal is to train a model $h$ that appraises the house price as $\hat{y_i}$ using the provided input defined above. The model function is defined as:

\begin{equation}
    h(\mathbf{f}_i, loc^x_i, loc^y_i; GEO^{POI}_{1, ..., N_P}, GEO^{ROAD}_{1, ..., N_R}) \rightarrow \hat{y}_i
\end{equation}


\section{Method}

\subsection{POI Feature Extraction}

Using the $loc^x_i$ and $loc^y_i$ for each house, we extract the POI feature by calculating the proximity to each POI geometry $GEO^{POI}_{1, ..., N_P}$ using the gaussian filter based on distance measure as
$    \mathbf{p}_{i} = [\text{Prox}_{i,\tau}] \quad \forall \tau \in \{1,...,N_P\}$, where $\text{Prox}_{i,\tau} = \exp\left(-{{ \text{Dist}_{i,\tau} ^2}} / {{2\beta^2}}\right)$.
We calculate the Euclidian distance of a $i$-th house to closest $\tau$-th geometry $\text{Dist}_{i,\tau}$ using a GIS software\footnote{GeoPandas (\url{https://geopandas.org/}) without CRS transformation (WGS84).}, and experimentally determine the appropriate $\beta$ value in Sec.~\ref{sec:beta-study}.

\subsection{Areal Embedding} \label{areal-embedding}

We propose two primary approaches to leverage areal embedding for the location of houses, diverging from the conventional use of latitude and longitude coordinates as extra house attributes as described in \cite{ASI}. Our method involves dividing an area into $M = M_x \times M_y$ grids, each comprising areas denoted as $\mathcal{A}_{1,...,M}$. For each of these areas, we construct a $D$-dimensional embedding. Each house corresponds to an area, enabling us to later use the respective areal embedding to infer its price in that specific location.

\subsubsection{2D Positional Encoding}

An initial avenue we can explore involves the application of sinusoidal 2D positional encoding, inspired by the work of \cite{pe2d}, as an alternative to using raw latitude and longitude coordinates. 
By incorporating this spatial information, the model can discern the correlation between house prices and location tendencies, resulting in an embedding in $\mathbb{R}^{M_x \times M_y \times D}$.

\subsubsection{Node2Vec Embedding}

In this method, we consider each line geometry of the $r$-th road, denoted as $GEO^{ROAD}_{r}$. We assume that the areas traversed by this road can be represented by the set $\mathcal{A}^{(r)} \subset \mathcal{A}_{1,...,M}$. Assuming interconnectedness between these areas, we count each ordered pair $(u, v) \in \mathcal{A}^{(r)} \times \mathcal{A}^{(r)}$, where $u \neq v$, as one connection. By applying this process to all $N_R$ roads, we construct the adjacency matrix $Adj \in \mathbb{N}^{M \times M}$, where $Adj_{[i,j]}$ denotes the number of roads connecting $\mathcal{A}_i$ and $\mathcal{A}_j$. Subsequently, we employ Node2Vec~\cite{node2vec} with $Adj$ to learn areal embeddings and visualize as Fig.~\ref{fig:areal-embedding}. Note, that the normalized weights of $Adj$ can be considered as the probability of random walks between areas, aligning with the fundamental concept of Node2vec.

\begin{figure}[t]
    \centering
    \includegraphics[width=\textwidth]{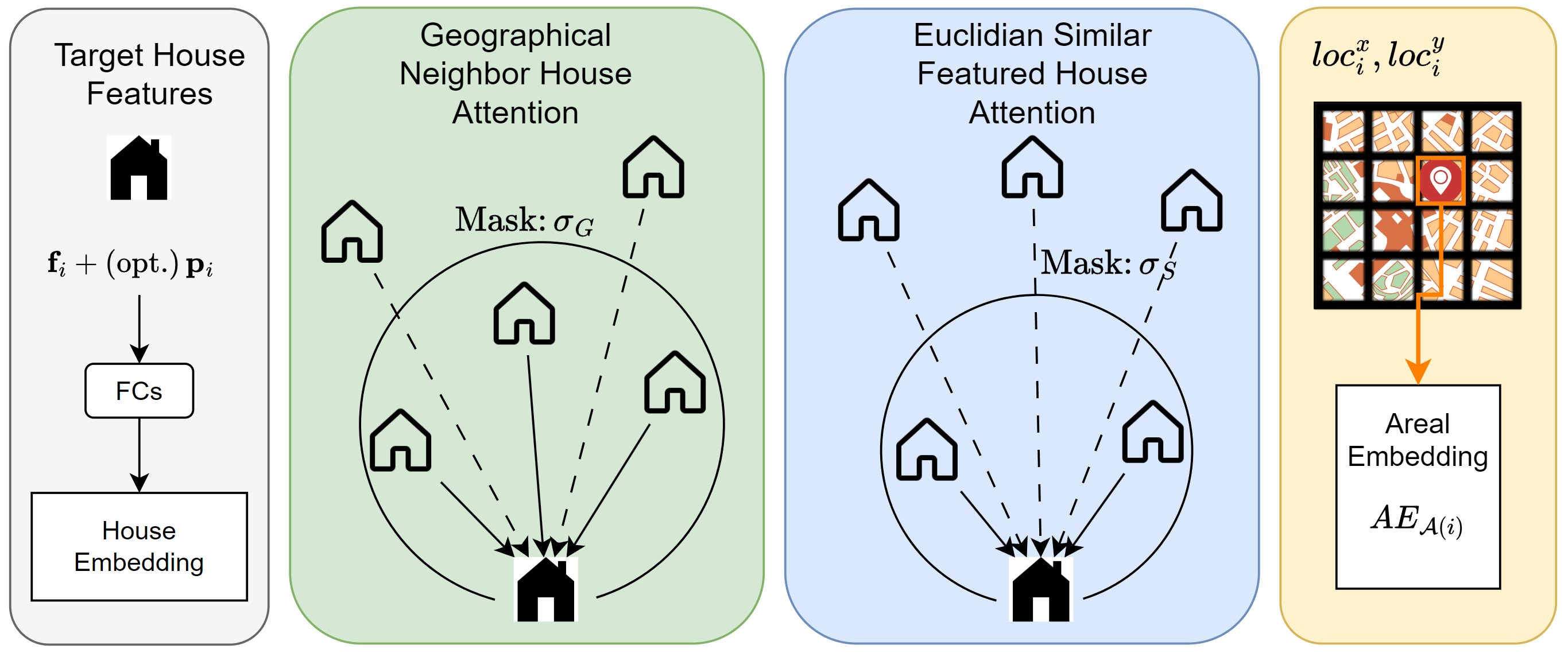}
    \caption{Proposed model architecture of AMMASI before the final regression layer.}
    \label{fig:ammasi-architecture}
\end{figure}

\subsection{Deep Neural Network (AMMASI)}

We introduce \textbf{A}real embedding-enabled \textbf{M}asked \textbf{M}ultihead \textbf{A}ttention-based \textbf{S}patial \textbf{I}nterpolation (AMMASI) for house price prediction. Our proposed model builds upon the foundational framework of ASI\cite{ASI}, specifically leveraging geographical neighbor houses and similar-featured houses. The comprehensive architecture is illustrated in Fig.~\ref{fig:ammasi-architecture}.

\subsubsection{House Feature Embedding Layer}

To encode the house features $\mathbf{f}_i$, we employ two stacked fully connected layers (FCs), resulting in $D$-dimensional embedding. Optionally, POI features $\mathbf{p}_i$ can be concatenated to $\mathbf{f}_i$ to enrich the house features before ingested to FCs.

\subsubsection{Masked Multi-head Attention on Reference Houses}

Similar to the architecture proposed in \cite{ASI}, our model employs a house attention mechanism operating on two types of house knowledge. First, we leverage information from geographically nearby houses and their prices, as they offer insights into the surrounding area. Secondly, we consider houses with similar features regardless of geographical distances and their corresponding prices. 
While building upon the foundational framework of ASI, we have identified a scalability limitation in implementing the attention mechanism, particularly when using the widely adopted dot product-based approach. To overcome this challenge, we enhance our model by introducing a masked multi-head attention mechanism tailored specifically for these two types of house attention.

Let $\textbf{Gidx}^{(i)} = \{\text{Gidx}^{(i)}_1, ..., \text{Gidx}^{(i)}_{N_G}\}$ and $\textbf{GDist}^{(i)} = \{\text{GDist}^{(i)}_1, ... , \text{GDist}^{(i)}_{N_G}\}$ represent the indexes and geographical distances of geographically neighboring houses based on location. Additionally, let $\textbf{Sidx}^{(i)} = \{\text{Sidx}^{(i)}_1, ..., \text{Sidx}^{(i)}_{N_S}\}$ and $\textbf{SDist}^{(i)} = \{\text{SDist}^{(i)}_1, ..., \text{SDist}^{(i)}_{N_S}\}$ denote the indexes and distances of houses with similar features, measured by Euclidean distances of house features. Here, $N_G$ and $N_S$ are the numbers of houses for attention of each module. As we leverage the same index and distance-measured legacy dataset provided by the authors of ASI, readers can readily refer to their definition in \cite{ASI} for further details. Note, that the reference houses are found among the houses that only exist in the training dataset, that is $(\textbf{Gidx}^{(i)} \cup \textbf{Sidx}^{(i)}) \subset \text{TrainDataset}$. Therefore, we can look up their house attributes and prices during the inference phase. Let concatenation of $i$-th house attributes with its price as $\textbf{f}^{\star}_i$. Then we can conduct the attention mechanism as follows:

\begin{align}
    emb^{(query)} &= f_q(\mathbf{f}_i) \in \mathbb{R}^{1 \times d} \\
    emb^{(key)} &= f_k([\mathbf{f}^{\star}_{\text{Gidx}^{(i)}_1}, ..., \mathbf{f}^{\star}_{\text{Gidx}^{(i)}_{N_G}}]) \in \mathbb{R}^{N_G \times d} \\
    emb^{(value)} &= f_v([\mathbf{f}^{\star}_{\text{Gidx}^{(i)}_1}, ..., \mathbf{f}^{\star}_{\text{Gidx}^{(i)}_{N_G}}]) \in \mathbb{R}^{N_G \times d} 
\end{align}

Here, $f_q, f_k, f_v$ are two-stacked FCs with ELU activation. 

\begin{equation}  
    {score}_j = \frac{\exp\left({\langle}  emb^{(query)}, emb^{(key)}_j{\rangle} / \sqrt{d} + Mask^{(i)}_j\right)}
    {\sum_{k=1}^{N_G} \exp\left({\langle}  emb^{(query)}, emb^{(key)}_k  {\rangle} / \sqrt{d} + Mask^{(i)}_k\right)} \in \mathbb{R}^{1 \times N_G}
\end{equation}

where ${\langle} \bullet, \bullet {\rangle}$ denotes inner product operator, and $Mask^{(i)}_j = -\infty$ if $\text{GDist}^{(i)}_{j} > \sigma_G$ else 0. Finally, we leverage $output_G = \sum_{j=1}^{N_G}{{score}_j \cdot {emb}^{(value)}_j} \in \mathbb{R}^{1 \times d}$ as our final output embedding. Suppose there are $K$ different attention heads that conduct the same function, then we can introduce $K$-attention head as $Output_G = f_G \left( \Big\|_{m}^{1,...,K} output_G^{(m)}  \right) $, where $||$ denotes the concatenation and $f_G$ as two-stacked FCs. We apply attention similarly to houses with Euclidean similar features, yielding $Output_S$ as a result. The values of $\sigma_G$ and $\sigma_S$ are determined empirically, and their discussion will be elaborated in Section \ref{sigma-study}.

\subsubsection{Areal Embedding Lookup}

Every house is associated with a specific area, denoted as $\mathcal{A}(i)$, where the location of the $i$-th house resides. Leveraging the pre-trained embedding values for each area introduced in Sec.~\ref{areal-embedding}, we incorporate areal embeddings before the final output layer in our model.

\subsubsection{Final Output Layer}

In the concluding steps, we concatenate the previously mentioned embeddings, including (1) house feature embedding, (2) attention output with geographically neighboring houses, (3) attention output with similar-featured houses, and (4) areal embedding, resulting in a $4 D$-dimensional embedding. Subsequently, we apply a final output dense layer, consisting of a two-stacked FCs with ELU activation, to generate the final output.

\subsubsection{Objective Function}

We train our model based on $L(\theta) = \frac{1}{n} \sum_{i=1}^{n} \left| log(y_i) - log(\hat{y}_i) \right|$ which is MAE of log value of the house prices.

\section{Evaluation Settings}

\subsection{Dataset}

We utilize four real estate datasets, namely FC, KC, SP, and POA, published by \cite{ASI}\footnote{\url{https://github.com/darniton/ASI}}. These datasets represent King County (KC) in Washington State, USA; Fayette County (FC) in Kentucky, USA; São Paulo (SP), Brazil; and Porto Alegre (POA), Brazil. 
We leverage the same \textbf{Gidx}, \textbf{GDist}, \textbf{Sidx}, \textbf{SDist} which are the indexes and the distance for the referencing houses for attention as the original paper for a fair comparison between ASI and AMMASI in the experiment.
Furthermore, we gather the geometry information for 15 types of OpenStreetMap (OSM)\footnote{\url{http://www.openstreetmap.org/}} points of interest (POI), as depicted in the columns of Fig.~\ref{fig:poi-study}.
We also collect road network data from OSM as well.
We applied the processed POI features $\mathbf{p}_i$ on both ASI and AMMASI\footnote{That is, we do not leverage the POI features in \cite{ASI} as it was not publicly available.}.



\subsection{Parameters}

\begin{figure}[t]
\centering
\includegraphics[width=\textwidth]{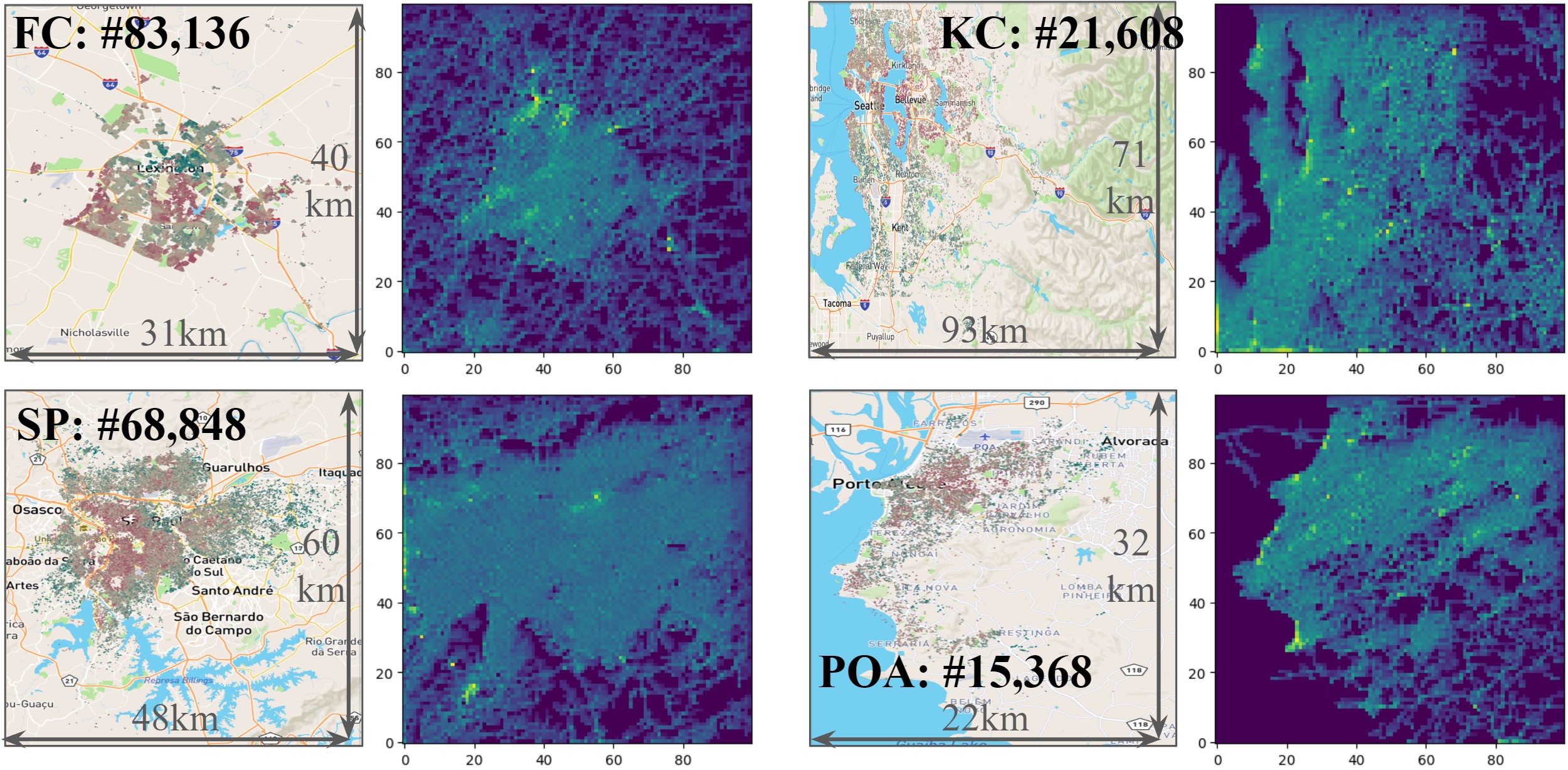}
\caption{Visualization of house prices (green $<$ red, with the house count \#) and Node2Vec areal vector magnitudes ($M_x \times M_y$ = 100 $\times$ 100).}
\label{fig:areal-embedding}
\end{figure}

We have chosen an embedding dimension of $D = 64$ for the house feature embedding, the two attention embedding outputs, and the areal embedding. For areal embedding, we split the region into $100 \times 100$ grid, resulting in $M=10,000$, and also illustrated in Fig.~\ref{fig:areal-embedding}. In the case of two-stacked FCs, we utilize a $D$-dimensional (c.f. $d$-dimensional for $f_q, f_k, f_v$) output of each layer. The first layer employs the ELU activation function, while the second layer has no activation. 
The dataset is split into training, validation, and test sets with a ratio of 0.72, 0.08, and 0.2, respectively as the same setting of \cite{ASI}.
For the multi-head attention, we have set $d = 8$ and $K = 8$. We leverage Adam optimizers, batch size of 250, initial learning rate of 0.008, conduct early stopping on 10 patients, and reduce the learning rate to 1/10 after 5 patients.

\subsection{Error Measures}

We evaluate the performance of the models by using three different error measures\footnote{\cite{ASI} incorrectly referred to MdAPE as MAPE, unlike their code implementation.} described as $\text{MALE} = \frac{1}{n} \sum_{i=1}^{n} \left| log(y_i) - log(\hat{y}_i) \right|$, $\text{RMSE} = \sqrt{\frac{1}{n} \sum_{i=1}^{n} \left( y_i - \hat{y}_i \right)^2}$, and $\text{MdAPE} = \text{median}\left(\frac{\left| y_1 - \hat{y}_1 \right|}{\left| y_1 \right|}, \frac{\left| y_2 - \hat{y}_2 \right|}{\left| y_2 \right|}, \ldots, \frac{\left| y_n - \hat{y}_n \right|}{\left| y_n \right|} \right)$.

\section{Results}

\subsection{POI Feature Significance}\label{sec:beta-study}

\subsubsection{Gaussian parameter ($\beta$) for POI proximity feature}

\begin{figure}
\centering
\vspace{-5mm}
\subfloat[Sample proximity heatmap (green $<$ red) of each house to a POI (leisure-golf\_course) on FC.\label{fig:sample-poi-heatmap}]{\includegraphics[width=0.42\textwidth]{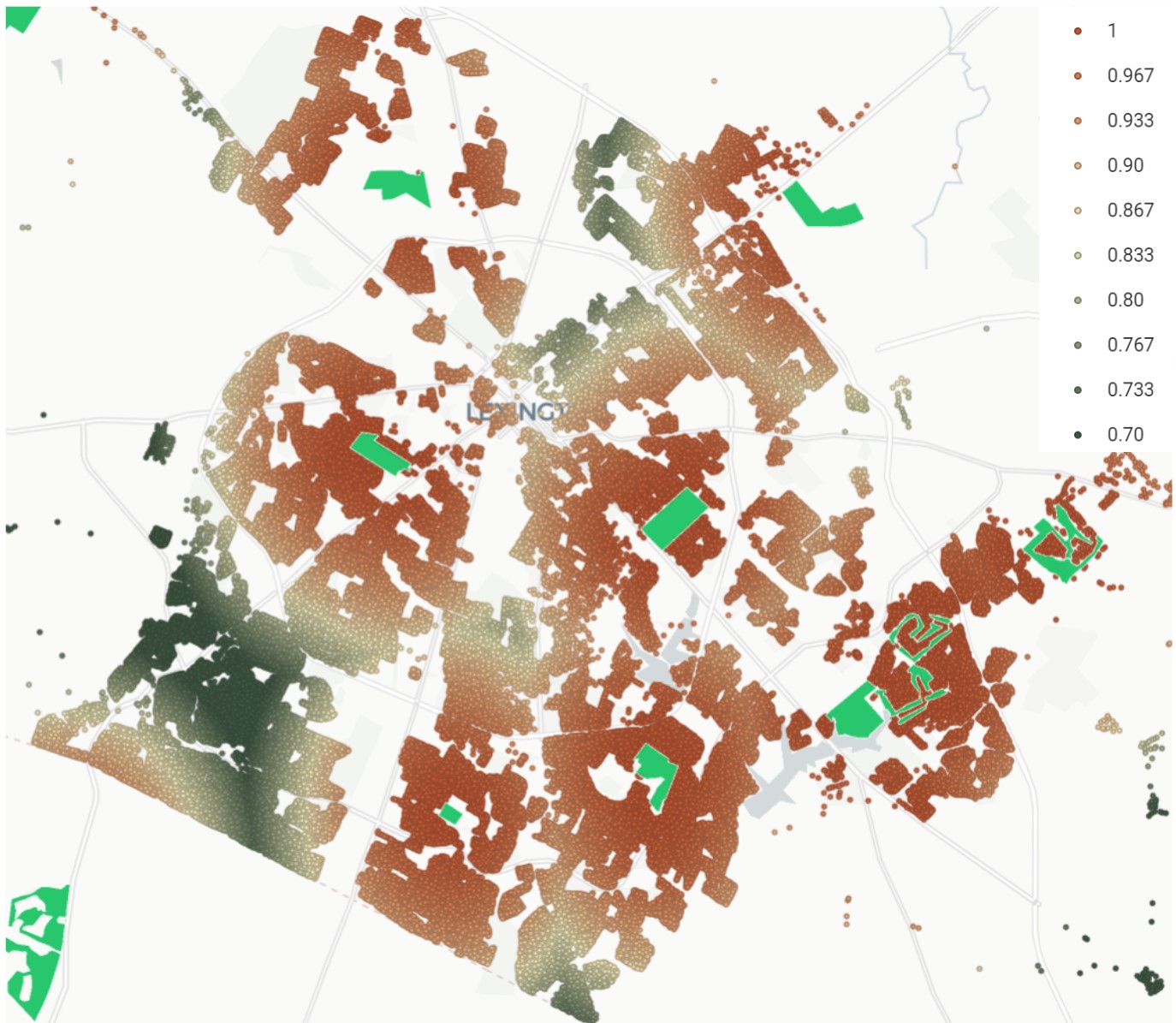}}
\hspace{2mm}
\subfloat[R-squared value of linear regression with varying $\beta$: (best at FC: 0.045, KC: 0.035, SP: 0.020, POA: 0.025)\label{fig:beta-study}]{\includegraphics[width=0.51\textwidth]{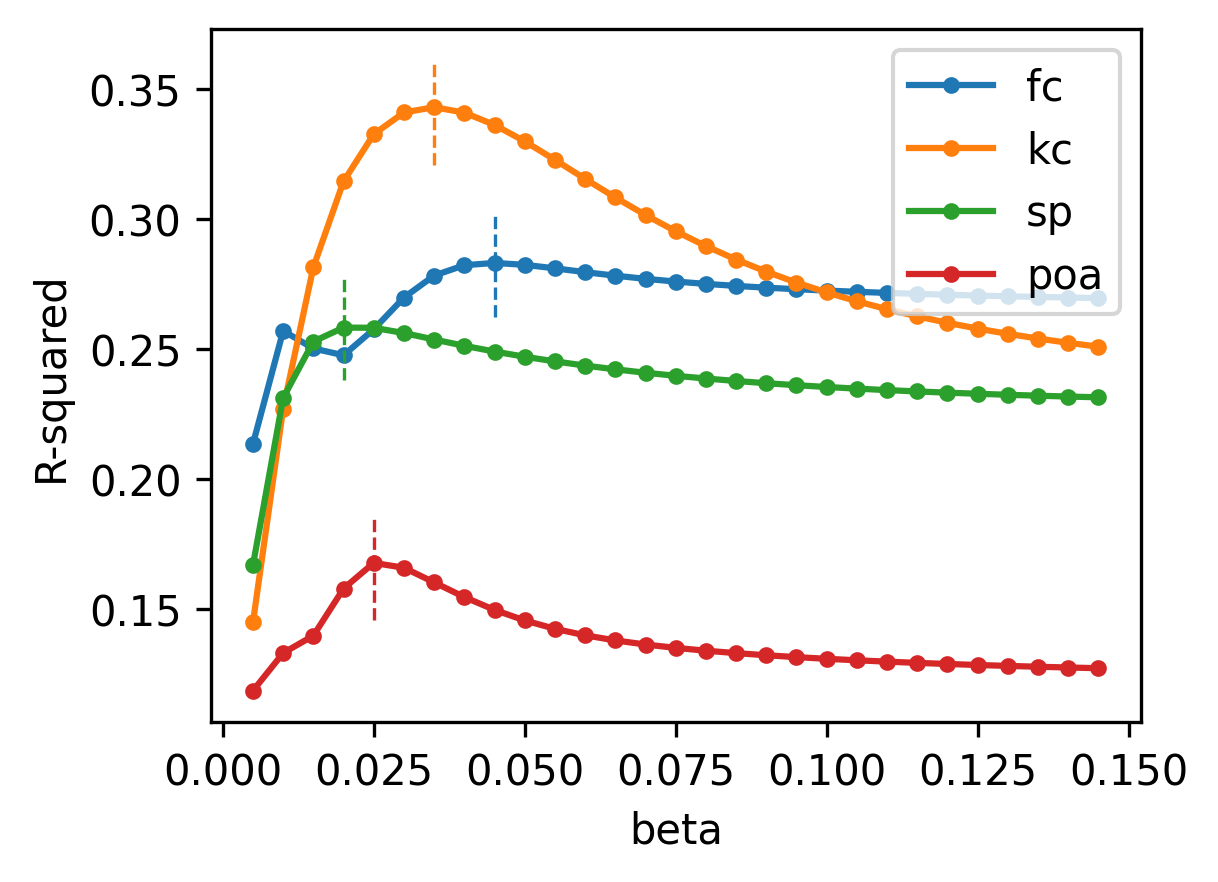}}
\caption{Proximity-based POI feature extraction.}
\vspace{-5mm}
\end{figure}

In Fig.~\ref{fig:sample-poi-heatmap} shows a heatmap of proximity to a specific POI type.
When calculating proximity, $\beta$ decides how distant POIs will be assumed to be considered important, as high $\beta$ value causes the proximity to consider POIs at greater distances as important.
Although, it is true that $\beta$ for each POI might be different, however, in this study we find the most optimal beta for each dataset region.
In Fig.~\ref{fig:beta-study}, we empirically find optimal $\beta$ as $\beta_\text{FC}= 0.045$, $\beta_\text{KC}= 0.035$, $\beta_\text{SP}= 0.020$, and $\beta_\text{POA}= 0.025$ which shows highest $R^2$ values in linear regression with $\mathbf{p}_i$ and $y_i$.

\subsubsection{Impact of different types of POIs.}

\begin{figure}[h]
\centering
\includegraphics[width=\textwidth]{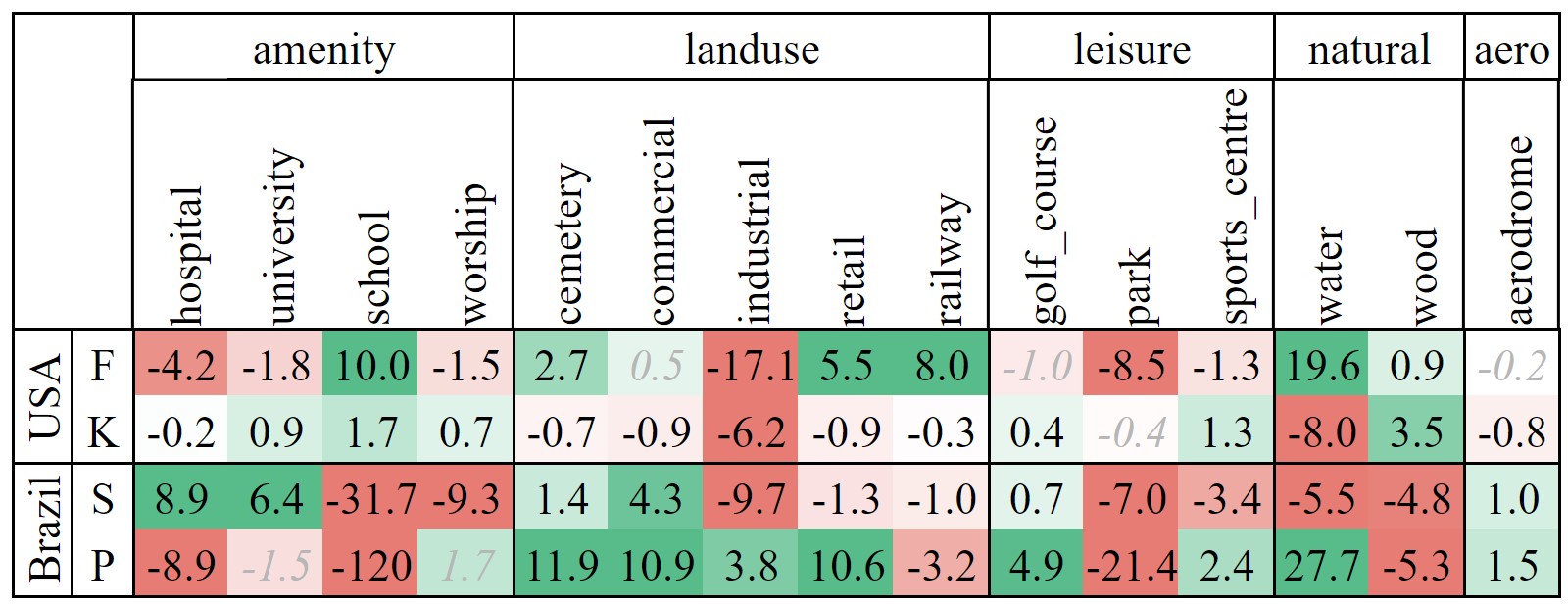}
\caption{Coefficients from linear regression for each POI proximity. Non-significant values (p-value $>$ 0.05) are highlighted in gray.}
\label{fig:poi-study}
\vspace{-6mm}
\end{figure}

Fig.~\ref{fig:poi-study} illustrates the coefficients derived from ordinary least squares regression for $\mathbf{p}_i$ and $y_i$, excluding $\mathbf{f}_i$ to emphasize the generalized trend of POI proximity and its impact on house prices. 
Notably, industrial land use demonstrates a negative influence on house prices. 
The impact of waterfront proximity is pronounced in FC and POA datasets, as also indicated by the Fig.~\ref{fig:areal-embedding} heatmap. 
However, these observations often deviate from conventional wisdom. 
Surprisingly, parks do not show a significant enhancement in house values in any region. 
Additionally, commercial areas particularly have a positive influence in Brazil. 
Moreover, while the presence of schools positively influences in the USA, it exerts a notably strong negative influence in Brazil.

The results can be interpreted in two ways: Firstly, the observed heterogeneity in the impact of features may be attributed to regional variations in culture, policies, and other human factors. Secondly, insufficient data engineering may play a role, particularly in the filtering process. Even among POIs of the same type, disparities in impact exist, suggesting a need for more nuanced approaches. Additionally, the varied application of $\beta$ values for each POI type may contribute to the observed deviations in impact.

\begin{table}[t] 
\caption{Performance comparison. HA: House Attribute, HA+P: HA + POI. * denotes the model leverages neighbor and similar house attention. Areal Embedding of AMMASI -- FC \& SP: Node2Vec, KC \& POA: sinusoidal.}
\centering
\begin{tabular}{cccccccccccc}
\toprule    
    & Model   & \multicolumn{2}{c}{LR} & \multicolumn{2}{c}{RF} & \multicolumn{2}{c}{XGBoost} & \multicolumn{2}{c}{ASI*} & \multicolumn{2}{c}{AMMASI*} \\ \cmidrule(l){3-4}  \cmidrule(l){5-6} \cmidrule(l){7-8} \cmidrule(l){9-10} \cmidrule(l){11-12}
    & Feature & HA         & HA+P    & HA         & HA+P    & HA           & HA+P       & HA         & HA+P     & HA         & HA+P     \\ \midrule

F & MALE    & 0.228  & 0.210  & 0.197  & 0.113  & 0.184   & 0.117  & 0.098  & 0.097          & 0.097          & \textbf{0.096}  \\
  & RMSE    & 52084  & 48716  & 41791  & 27035  & 38736   & 26087  & 23177  & \textbf{22678} & 23451          & 22918           \\
  & MAPE    & 15.60  & 14.58  & 13.17  & 7.14   & 12.59   & 8.00   & 6.49   & 6.29           & \textbf{6.09}  & 6.12            \\ \midrule
K & MALE    & 0.245  & 0.189  & 0.161  & 0.141  & 0.132   & 0.129  & 0.113  & 0.118          & 0.113          & \textbf{0.112}  \\
  & RMSE    & 246865 & 198921 & 174138 & 161233 & 145130  & 138499 & 115763 & 133543         & 108748         & \textbf{104582} \\
  & MAPE    & 20.19  & 14.90  & 11.32  & 9.87   & 9.64    & 9.39   & 8.00   & 8.40           & \textbf{7.82}  & 7.95            \\ \midrule
S & MALE    & 0.271  & 0.227  & 0.243  & 0.153  & 0.241   & 0.160  & 0.135  & 0.144          & 0.135          & \textbf{0.133}  \\
  & RMSE    & 267317 & 235776 & 250505 & 170105 & 244170  & 172915 & 155585 & 162520         & 156099         & \textbf{154483} \\
  & MAPE    & 23.12  & 18.91  & 19.72  & 11.34  & 19.73   & 12.60  & 9.80   & 10.82          & 9.56           & \textbf{9.51}   \\ \midrule
P & MALE    & 0.271  & 0.236  & 0.251  & 0.163  & 0.246   & 0.172  & 0.143  & 0.151          & \textbf{0.139} & 0.142           \\
  & RMSE    & 154878 & 140237 & 147384 & 102851 & 139401  & 104024 & 94492  & 98615          & \textbf{93246} & 94522           \\
  & MAPE    & 23.26  & 20.33  & 20.65  & 11.40  & 20.60   & 13.98  & 9.58   & 10.89          & \textbf{8.89}  & 9.36            \\ \bottomrule
  
\label{tab:performance-comparison}
\end{tabular}
\vspace{-6mm}
\end{table}

\subsection{Performance Comparision} \label{sec:performance-comparison}

Table~\ref{tab:performance-comparison} presents a performance comparison relative to baseline models. The baseline models under consideration include linear regression (LR), random forest (RF), XGBoost, and ASI\cite{ASI}. Here, HA refers to using only House Attribute features, while HA+P indicates the inclusion of POI features with HA.

Firstly, LR, RF, and XGBoost models demonstrate substantial improvement in predicting house prices when incorporating POI features along with house attributes. This underscores the significance of utilizing POI information, aligning with findings in existing research.

Moreover, AMMASI consistently outperforms ASI. 
The optimal choice between HA and HA+P for ASI in each dataset resulted in an average improvement of 0.34 MAPE on AMMASI.
However, ASI and AMMASI show only marginal performance gains or even degradation when incorporating POI features.
ASI shows a higher MAPE of 0.63 when incorporating POI, whereas AMMASI shows a higher MAPE of 0.15 upon the addition of POI, based on the average results across four datasets.
This can be attributed to the fact that ASI and AMMASI have access to information on prices of neighboring or similarly featured houses, minimizing the need for additional knowledge gained from including POI data. Furthermore, ASI frequently demonstrates a decline in performance with the addition of POI features, particularly in instances such as KC, SP, and POA. This suggests that the existing methods of integrating POI may lead to overfitting when simply concatenated with house attributes. In contrast, AMMASI does not demonstrate notable performance degradation even with the inclusion of the POI feature; in fact, an improvement is observed in terms of RMSE rather than MAPE. This may be attributed to enhancements in the attention mechanism and the incorporation of areal embeddings.

\subsection{Parameter and Ablation Test}

\subsubsection{Threshold on masked attention ($\sigma_G$, $\sigma_S$).} \label{sigma-study}

\begin{figure}[t]
\centering
\subfloat[FC ($\sigma_S$ = 0.02)\label{fig:ab1}]{\includegraphics[width=0.49\textwidth]{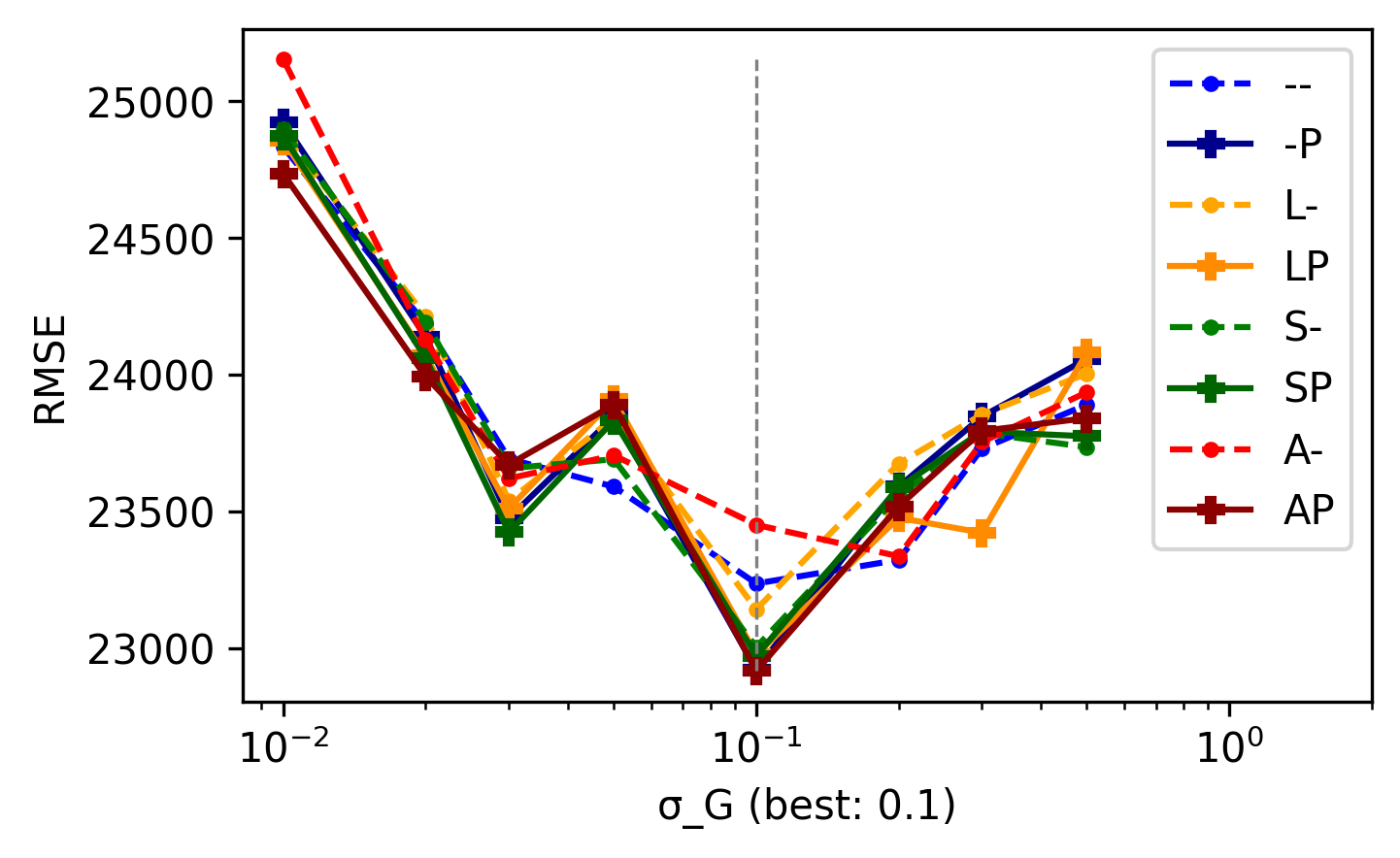}}
\subfloat[KC ($\sigma_S$ = 0.03)\label{fig:ab2}]{\includegraphics[width=0.49\textwidth]{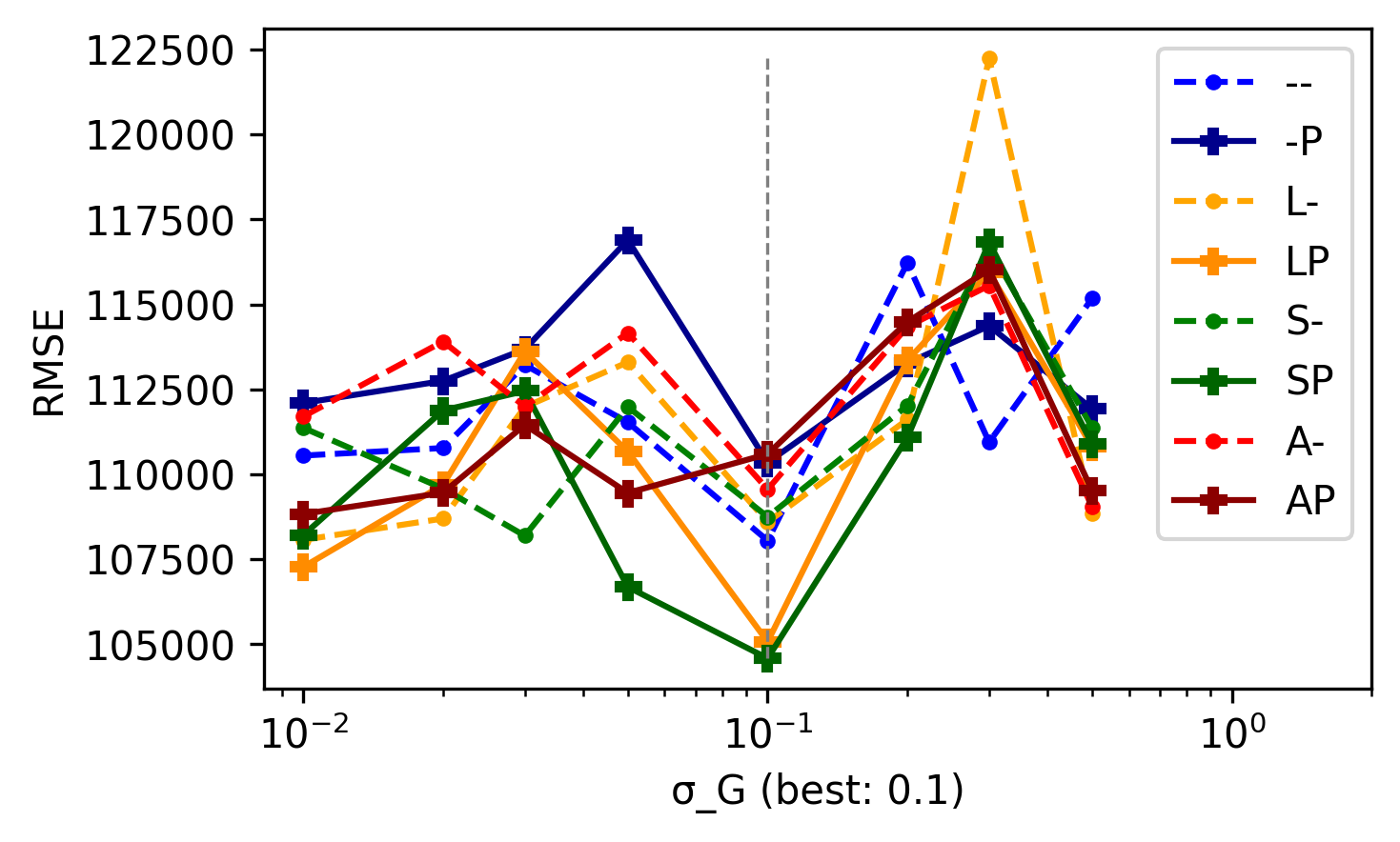}}
\vskip\baselineskip
\subfloat[SP ($\sigma_S$ = 0.01)\label{fig:ab3}]{\includegraphics[width=0.49\textwidth]{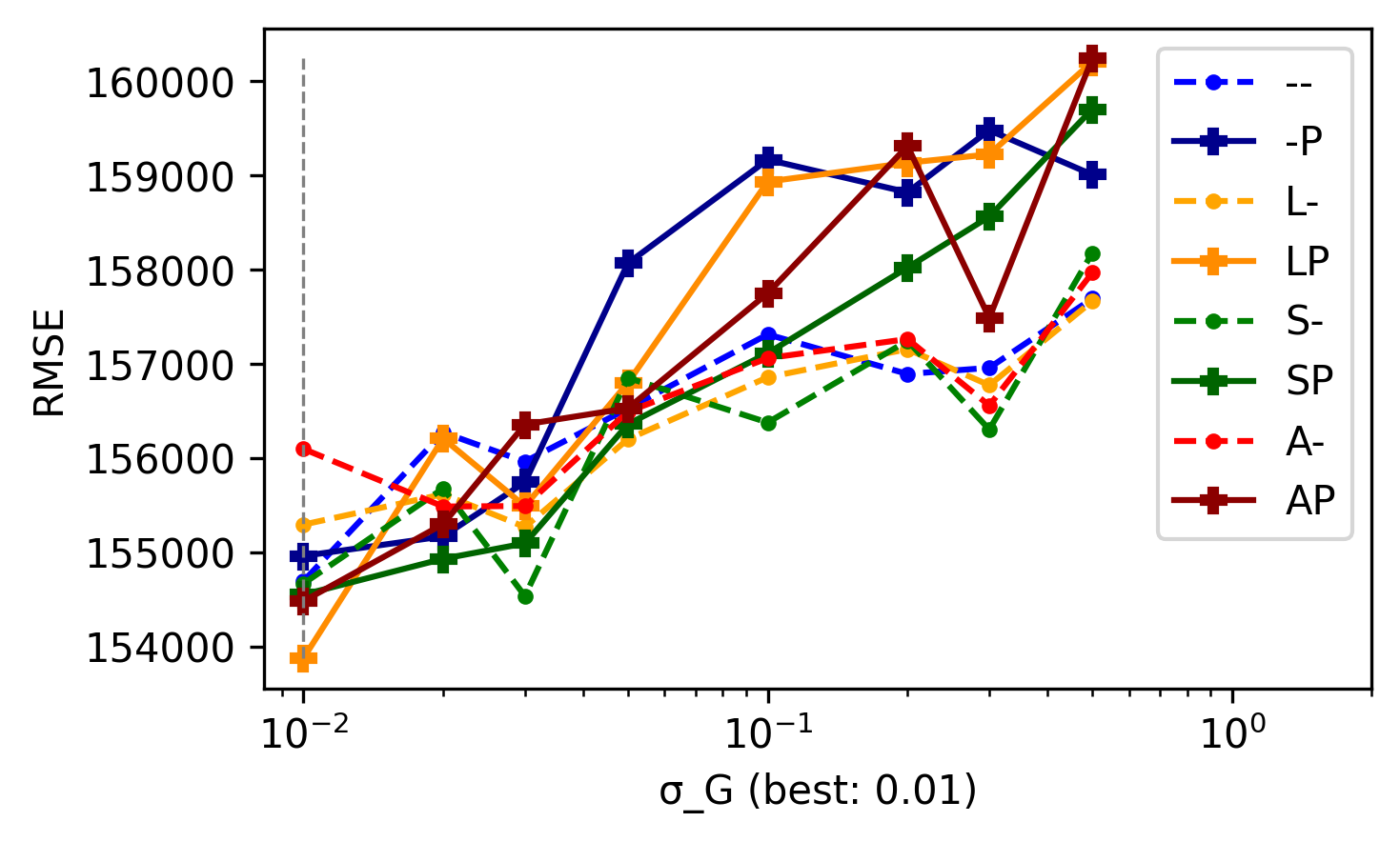}}
\subfloat[POA ($\sigma_S$ = 0.01)\label{fig:ab4}]{\includegraphics[width=0.49\textwidth]{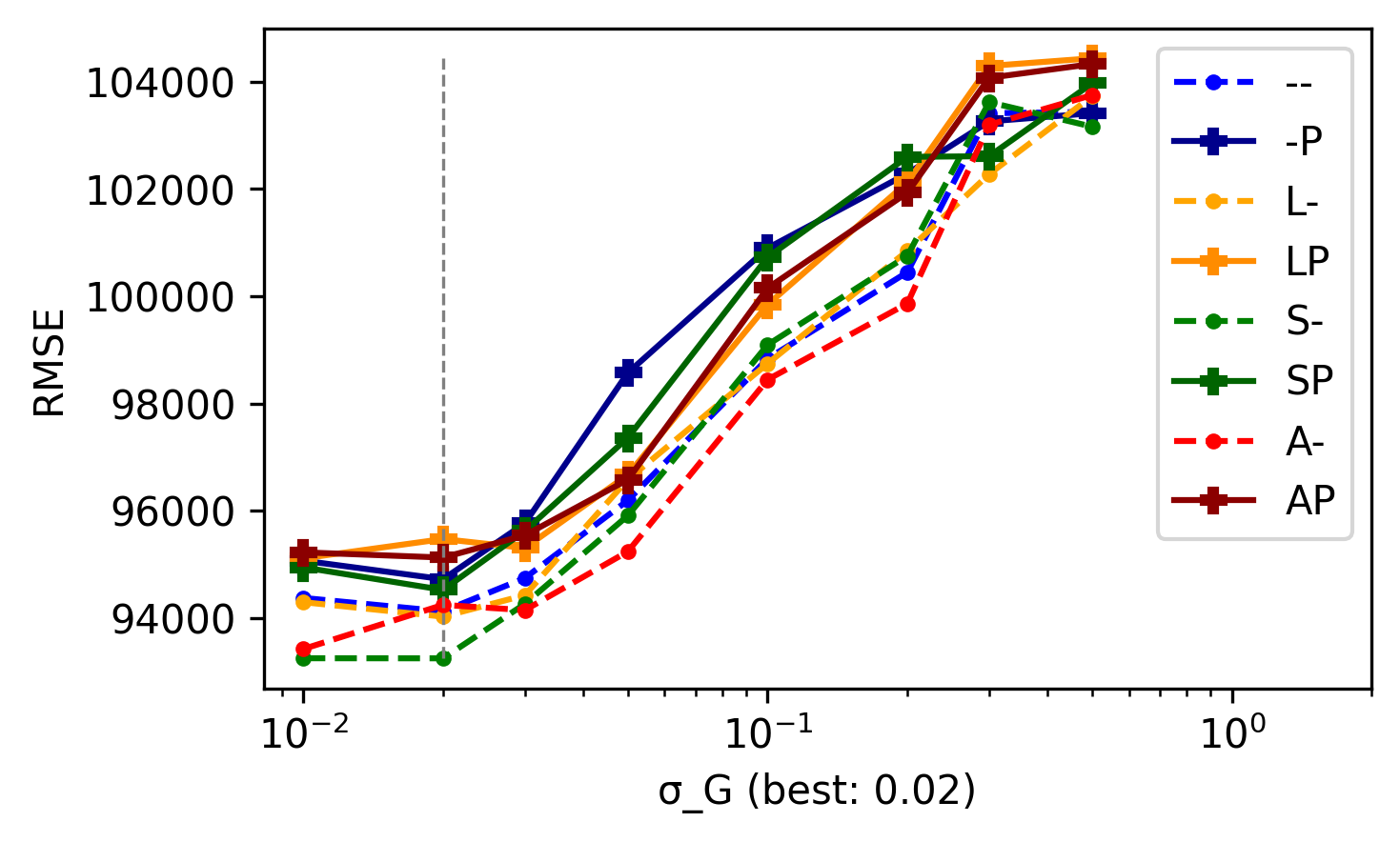}}
\caption{Study on $\sigma_G$ (with experimentally best $\sigma_S$) and an ablation study.}
\label{fig:sigma-study}
\vspace{-5mm}
\end{figure}


We empirically determine the optimal pair of $\sigma_G$ and $\sigma_S$\footnote{Testing every case of $(\sigma_G, \sigma_S) \in \{0.01, 0.02, 0.03, 0.05, 0.1, 0.2, 0.3, 0.5\}^2$}. Initially, we identify the most effective $\sigma_S$ values, wherein our model with areal embedding exhibits the least RMSE error. The parameter $\sigma_S$ influences how attention is assigned to houses with similar features, potentially influenced by Euclidean distance measures.

Simultaneously, Fig.~\ref{fig:sigma-study} illustrates the impact of $\sigma_G$ on model performance. The parameter $\sigma_G$ represents the lat/lon-based distance threshold for attention. The varying values of $\sigma_G$ and $\sigma_S$ for each region highlight the dataset's heterogeneity across different geographic areas.

\subsubsection{Comparison on Areal Embedding}

The analysis presented in Fig.~\ref{fig:sigma-study} investigates the impact of feature inclusion or exclusion, with a specific focus on areal embedding and POI. Four primary approaches are considered: (--) None, (L) lat/lon as additional attributes, (S) sinusoidal embedding, and (A) Node2Vec embedding. Furthermore, for each case, we explore the scenarios with or without the inclusion of POIs (--, P), resulting in a total of eight cases.

The results indicate that our areal embedding approach is notably effective, especially on SP:S-- and POA:A--. Furthermore, POI features show effectiveness in the FC and KC scenarios but have a lesser impact on SP and POA. Plus, Node2vec does not promise performance improvement over simple 2D sinusoidal. We attribute these observations to inaccuracies in the recorded latitude and longitude of houses and the comparatively commercial-area-centered urban planning in Brazil compared to the USA. Furthermore, we believe that an improvement in the granularity of area embedding could provide nuanced differences in embedding for houses.

\section{Conclusion}

In conclusion, this paper addresses key challenges in real estate appraisal, emphasizing the integration of Points of Interest (POI) and the implementation of Areal Embedding for enhanced property valuation. The introduced AMMASI model surpasses current baselines by incorporating POI data and utilizing a simplified road network-based Areal Embedding approach with the masked multi-head attention. Our contributions include the introduction of AMMASI with publicly available code, and a more effective method for POI feature selection and geographical insights, collectively advancing real estate appraisal methodologies. Notably, areal embedding opens avenues for integrating diverse urban datasets, including census and geographic spatial features as an extra channel, suggesting exciting possibilities for future research endeavors.

%
%
\bibliographystyle{splncs04}
\bibliography{mybibliography.bib}

\begin{thebibliography}{10}
\providecommand{\url}[1]{\texttt{#1}}
\providecommand{\urlprefix}{URL }
\providecommand{\doi}[1]{https://doi.org/#1}

\bibitem{PDVM}
Bin, J., Gardiner, B., Li, E., Liu, Z.: Peer-dependence valuation model for
  real estate appraisal. Data-Enabled Discovery and Applications  \textbf{3},
  1--11 (2019)

\bibitem{de2018economic}
De~Nadai, M., Lepri, B.: The economic value of neighborhoods: Predicting real
  estate prices from the urban environment. In: 2018 IEEE 5th International
  Conference on Data Science and Advanced Analytics (DSAA) (2018)

\bibitem{DoRA}
Du, W.W., Wang, W.Y., Peng, W.C.: Dora: Domain-based self-supervised learning
  framework for low-resource real estate appraisal. In: Proc. of the ACM
  International Conference on Information and Knowledge Management (2023)

\bibitem{node2vec}
Grover, A., Leskovec, J.: node2vec: Scalable feature learning for networks. In:
  Proc. of the ACM SIGKDD international conference on Knowledge discovery and
  data mining. pp. 855--864 (2016)

\bibitem{LESR}
Jenkins, P., Farag, A., Wang, S., Li, Z.: Unsupervised representation learning
  of spatial data via multimodal embedding. In: Proc. of the ACM International
  Conference on Information and Knowledge Management (2019)

\bibitem{STRAP}
Lee, H., Jeong, H., Lee, B., Lee, K.D., Choo, J.: St-rap: A spatio-temporal
  framework for real estate appraisal. In: Proc. of the ACM International
  Conference on Information and Knowledge Management (2023)

\bibitem{ReGram}
Li, C., Wang, W., Du, W., Peng, W.: Look around! {A} neighbor relation graph
  learning framework for real estate appraisal. In: Proc. the AAAI Conference
  on Artificial Intelligence, MuFin Workshop (2022)

\bibitem{ottensmann2008urban}
Ottensmann, J.R., Payton, S., Man, J.: Urban location and housing prices within
  a hedonic model. Journal of Regional Analysis and Policy  \textbf{38}(1)
  (2008)

\bibitem{rosen1974hedonic}
Rosen, S.: Hedonic prices and implicit markets: product differentiation in pure
  competition. Journal of political economy  \textbf{82}(1),  34--55 (1974)

\bibitem{ASI}
Viana, D., Barbosa, L.: Attention-based spatial interpolation for house price
  prediction. In: Proc. of the International Conference on Advances in
  Geographic Information Systems (2021)

\bibitem{road2vec}
Wang, M.X., Lee, W.C., Fu, T.Y., Yu, G.: On representation learning for road
  networks. ACM Transactions on Intelligent Systems and Technology (TIST)
  \textbf{12}(1),  1--27 (2020)

\bibitem{pe2d}
Wang, Z., Liu, J.C.: Translating math formula images to latex sequences using
  deep neural networks with sequence-level training. International Journal on
  Document Analysis and Recognition (IJDAR)  \textbf{24}(1-2),  63--75 (2021)

\bibitem{Hex2vec}
Wo{\'z}niak, S., Szyma{\'n}ski, P.: Hex2vec: Context-aware embedding h3
  hexagons with openstreetmap tags. In: Proc. of the ACM SIGSPATIAL
  International Workshop on AI for Geographic Knowledge Discovery (2021)

\bibitem{xiao2017beijingPOI}
Xiao, Y., Chen, X., Li, Q., Yu, X., Chen, J., Guo, J.: Exploring determinants
  of housing prices in beijing: An enhanced hedonic regression with open access
  poi data. ISPRS International Journal of Geo-Information  \textbf{6}(11),
  ~358 (2017)

\bibitem{MugRep}
Zhang, W., Liu, H., Zha, L., Zhu, H., Liu, J., Dou, D., Xiong, H.: Mugrep: A
  multi-task hierarchical graph representation learning framework for real
  estate appraisal. In: Proc. of the ACM SIGKDD on Knowledge Discovery \& Data
  Mining (2021)

\end{thebibliography}
\end{document}